\documentclass{bmvc2k}

\usepackage{booktabs} 
\usepackage{amsmath}  
\usepackage{amssymb} 
\usepackage{multirow} 
\usepackage{siunitx} 
\usepackage{tabularx}
\usepackage{algorithm}
\usepackage{algpseudocode}
 
\newcommand{\best}[1]{\textcolor{red}{#1}}    % best-red
\newcommand{\second}[1]{\textcolor{blue}{#1}} % second_best-blue
\title{A Novel Local Focusing Mechanism for Deepfake Detection Generalization}

\addauthor{Mingliang Li}{limingliang@jxnu.edu.cn}{1}
\addauthor{Lin Yuanbo Wu}{Xiaoxian.wu9188@gmail.com}{2}
\addauthor{Changhong Liu}{liuch@jxnu.edu.cn}{1}
\addauthor{Hanxi Li}{lihanxi2001@foxmail.com}{1}

\addinstitution{
 Jiangxi Normal University,  China
}
\addinstitution{
 Swansea University, United Kingdom
}

\runninghead{Student, Prof, Collaborator}{BMVC Author Guidelines}

\begin{document}

\maketitle
\begin{abstract}
The rapid advancement of deepfake generation techniques has intensified the need for robust and generalizable detection methods. Existing approaches based on reconstruction learning typically leverage deep convolutional networks to extract differential features. However, these methods show poor generalization across object categories (e.g., from faces to cars) and generation domains (e.g., from GANs to Stable Diffusion), due to intrinsic limitations of deep CNNs. First, models trained on a specific category tend to overfit to semantic feature distributions, making them less transferable to other categories, especially as network depth increases. Second, Global Average Pooling (GAP) compresses critical local forgery cues into a single vector, thus discarding discriminative patterns vital for real-fake classification.

To address these issues, we propose a novel Local Focus Mechanism (LFM) that explicitly attends to discriminative local features for differentiating fake from real images. LFM integrates a Salience Network (SNet) with a task-specific Top-K Pooling (TKP) module to select the $K$ most informative local patterns. To mitigate potential overfitting introduced by Top-K pooling, we introduce two regularization techniques: Rank-Based Linear Dropout (RBLD) and Random-K Sampling (RKS), which enhance the model's robustness. LFM achieves a 3.7\% improvement in accuracy and a 2.8\% increase in average precision over the state-of-the-art Neighboring Pixel Relationships (NPR) method~\cite{tan2024rethinking}, while maintaining exceptional efficiency at 1789 FPS on a single NVIDIA A6000 GPU. Our approach sets a new benchmark for cross-domain deepfake detection. The source code are available in
  \url{https://github.com/lmlpy/LFM.git}
\end{abstract}

%-------------------------------------------------------------------------
\section{Introduction}
\label{sec:intro}
With the rapid development of image synthesis technologies, including GAN~\cite{goodfellow2014generative,karras2017progressive,karras2019style,Liu-ESWA-2025} and Stable Diffusion~\cite{ho2020denoising,rombach2022high,LIU-PRAI-2024}, the images generated have become more and more lifelike. There is no doubt that these technologies bring convenience to our daily lives, but the undetectability inherent in deepfake synthesis also causes non-negligible risks to our daily lives, such as identity forgery and false information dissemination. To cope with the potential threat from deepfake, developing a robust deepfake detection model is imperatively demanded. And these~\cite{tan2024rethinking,liu2024forgery,li2024improving} are some recent works on deepfake detection.

\begin{figure}[t]
    \centering
    \includegraphics[width=1\linewidth]{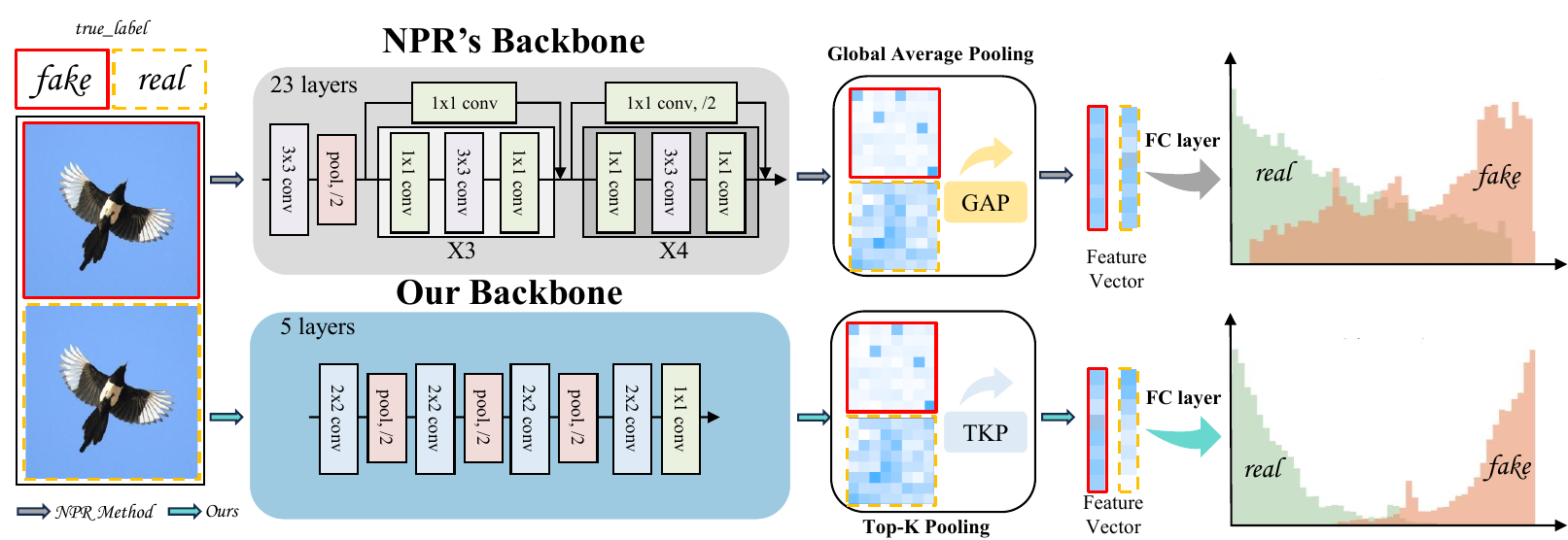}
    \caption{The differences between ours and the NPR~\cite{tan2024rethinking}. Our method employs a simpler network with task-specific TKP, in contrast to NPR's ResNet50 with GAP.}
    \label{fig1:differences}
\end{figure}

The main methods of deepfake detection are deep learning, which can be categorized into three groups according to their original feature extraction paradigms: time-space~\cite{rossler2019faceforensics++,ojha2023towards,tan2023learning}, frequency domain~\cite{durall2020watch,qian2020thinking,liu2024forgery}, and reconstruction-based methods~\cite{cao2022end,tan2024rethinking}. However, these methods usually employ deep CNNs to construct the classifier, which brings about two key issues. Firstly, training on a dataset of one category (e.g., cats) inevitably leads to fitting the distribution of semantic features extracted from the deep-layer networks. Consequently, the trained model fails to generalize effectively to datasets of another category (e.g., dogs), which is named the limitation of category generalization. Secondly, the GAP compresses critical local forgery cues into a single vector that tends to focus on those numerous yet non-discriminative forgery patterns and discard those sparse yet discriminative local forgery patterns. It inevitably leads to the model being coercively fitted to a data distribution generated by a generation source (e.g., GAN), making it fail to generalize effectively to datasets generated by another generation source (e.g., Stable Diffusion), which is named the limitation of generation source generalization.

In this paper, we propose a novel Local Focusing Mechanism (LFM) based on Neighboring Pixel Relationships (NPR)~\cite{tan2024rethinking} feature as input. We implement LFM efficiently via a Salience Network (SNet) combined with task-specific Top-K Pooling (TKP) strategy, which can solve these problems by limiting receptive fields and preferentially selecting the top-k most discriminative local forgery patterns. However, TKP only focuses on sparse yet discriminative local forgery patterns that reduces the difficulty of deepfake detection, which introduces additional risk of potential overfitting. Therefore, we propose Rank-Based Linear Dropout (RBLD) to reduce the model's reliance on the most discriminative local forgery patterns. In addition, we employ Random-K Sampling (RKS) to introduce a little disturbance to enhance model robustness. The difference between the NPR~\cite{tan2024rethinking} and our method is shown in Figure~\ref{fig1:differences}, where our method employs a simpler network combined with task-specific TKP, in contrast to NPR’s ResNet50 combined GAP, and our method has significantly higher discriminative power between fake and real images.

The main contributions of this paper include: 1) A local focusing mechanism is proposed to restrict receptive fields and preferentially select the top-k most discriminative local forgery patterns, which is implemented by SNet combined with TKP. 2)
The validation result from 28 generation sources shows that compared with NPR~\cite{tan2024rethinking}, the accuracy is increased from 92.2\% to 95.9\%, the average accuracy is increased from 95.8\% to 98.6\%, and it can reach 1789 FPS with the A6000 over other methods.
\section{Related work}

\noindent\textbf{Spatial-Temporal Methods:} Rossler et al.~\cite{rossler2019faceforensics++} created a large-scale dataset called FaceForensics++, which is used to train a binary classification model for deepfake detection. Subsequently, Franc et al. employed a simple image-trained Xception~\cite{chollet2017xception} to detect fake face images, while several studies~\cite{li2018ictu,haliassos2021lips} focus on specific areas to detect patterns of AI forgeries. In parallel, Chai et al.~\cite{chai2020makes} employed a limited receptive field to extract local features to find patches with forged patterns in forged images. Furthermore, Ju et al.~\cite{ju2022fusing} integrated global spatial information and local information features to train a two-branch model. Additionally, other research efforts~\cite{wang2021representative,wang2020cnn,chen2022self,ZhongICME-2025,CTVIS-2023,Wang-TMM-2025} have enhanced the detector's generalization of invisible generation sources by increasing the diversity of training. Recently, Tan et al.~\cite{tan2024rethinking} considered the process of down-sampling and up-sampling in the generated model, then proposed a general representation of the forgery patterns based on the relationship between adjacent pixels, achieving a better generalization effect.

\noindent\textbf{Frequency Domain Methods:} 
Given that GAN architecture relies heavily on extended operations, some studies~\cite{durall2020watch,frank2020leveraging,masi2020two} have deeply studied the influence of up-sampling on the entire image, taking the spectrum as a representation of up-sampling artifacts. Subsequently, F3Net~\cite{qian2020thinking} extracts different frequency components, performs frequency statistics between information in local image blocks of real and fake images, and then introduces the difference into face forgery detection. Furthermore, Add~\cite{woo2022add} proposed two Distillation modules for detecting deepfake of high compression, and BiHPF~\cite{jeong2022bihpf}  amplified the amplitude of the artifact through two high-pass filters. Recently, Liu et al. proposed that the pre-trained model of Fatformer~\cite{liu2024forgery} combined with CLIP-ViT has a newly high detection effect on GAN generation sources.

\noindent\textbf{Reconstruction-Based Methods:}
Firstly, these methods train the network model with real samples and directly identify the samples with significant reconstruction errors as deepfake products. Then, some studies~\cite{cao2022end,yan2024jrc,huang2020fakepolisher} proposed a deepfake detection framework that creatively uses reconstruction differences to represent forgery patterns and inputs them into a classifier for deepfake detection. Compared with the previous two kinds of methods, the method based on reconstruction learning can amplify the artifacts of forged images, which has greatly improved the performance of the detector. The NPR recently proposed by Tan et al.~\cite{tan2024rethinking} is also based on up-down sampling reconstruction and amplify of forgery patterns that improve the detection ability of the model.
\section{Methodology}

\subsection{Problem Definition}
Deepfake detection aims to capture unnatural artifacts introduced by generative models such as GAN and Diffusion Models during image synthesis, which need to construct a discriminant model capable of distinguishing between real and fake images. Given the input image $I_{(i,j)}^{(c)} \in \mathbb{R}^{H \times W \times 3}$ to be detected, where $(i,j)$ and $c$ represent the spatial position and the index-channel, respectively. Deepfake detection requires building a detector $M$ combining $\sigma(\cdot)$ as the activation function to predict the probability $P = \sigma \left( M \begin{pmatrix} I_{(i,j)}^{(c)} \end{pmatrix} \right)$ that an image is fake. For a deepfake detection task, the $X_{\text{test}} = \left\{ I_{1(i,j)}^{(c)}, \ldots, I_{n(i,j)}^{(c)} \right\}$ is a set of images to be tested, and we can predict the final label by trained $f\left(\sigma(M\left(I_{\left(i,j\right)}^{\left(c\right)}\right)\right):I_{\left(i,j\right)}^{\left(c\right)}\rightarrow\{0,1\}$, where 0 represents the real image, 1 represents the forged image, and $f$ is a decision function.

\subsection{The Method}
\begin{figure}[t]
    \centering
\includegraphics[width=1\linewidth]{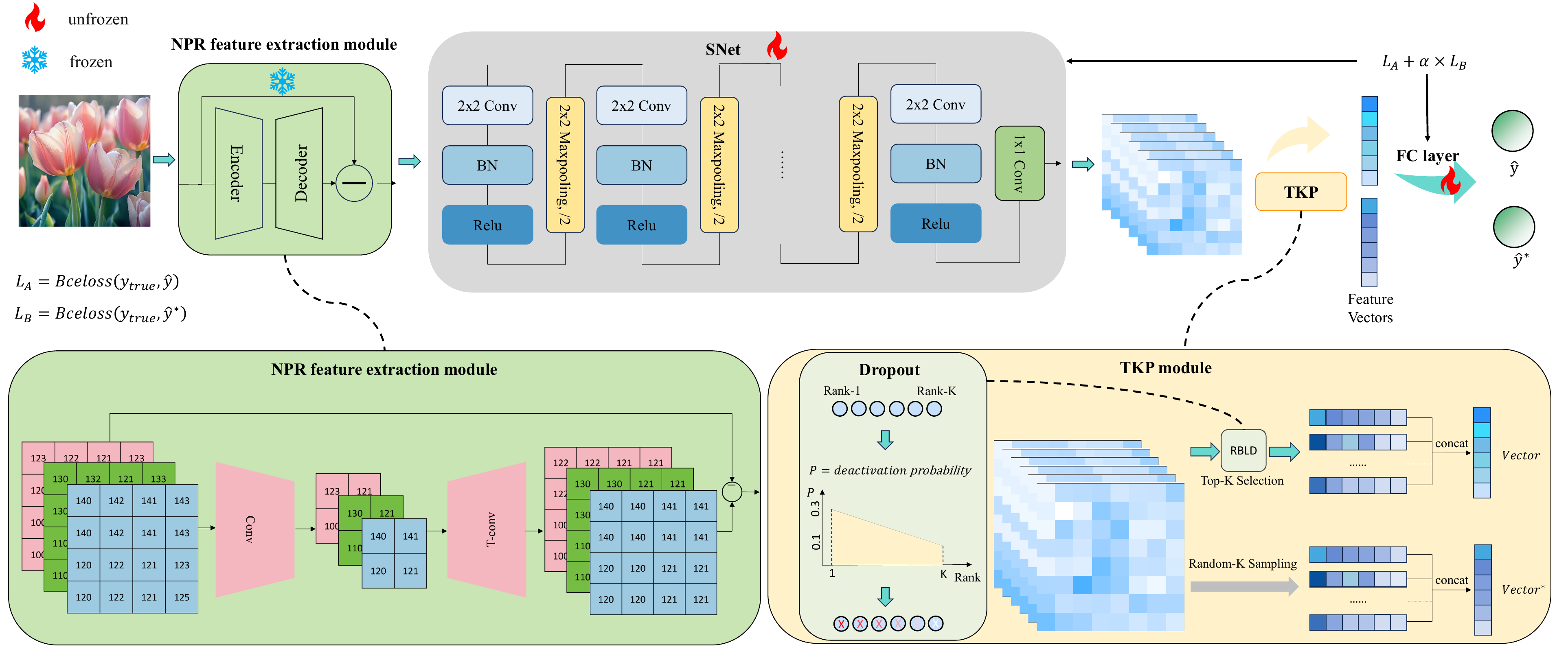}
    \caption{The overview of our method. The proposed LFM model consists of 3 sub-networks, namely the Neighboring Pixel Relationships (NPR), the simple yet efficient Salience Network (SNet) and Top-K-Pooling (TKP), respectively.}
    \label{fig2:view}
\end{figure}

As shown in Figure~\ref{fig2:view}, the proposed LFM model consists of 3 sub-networks, namely the Neighboring Pixel Relationships (NPR) for feature extraction, the  Salience Network (SNet) and Top-K Pooling (TKP).

\textbf{NPR based Feature Extraction:} We use the down-up-sampling reconstruction which can apply NPR feature~\cite{tan2024rethinking} to produce differential features. Assume each image input is $I_{(i,j)}^{(c)}\in  \mathbb{R}^{H\times W\times3}$, we down-up-sample by a series of convolution ($ \mathcal{C}_{\theta}$) and transpose convolution ($\mathcal{T}_{\theta}$) to yield the outputs $\mathbf{NPR}=I_{(i,j)}^{(c)} - \mathcal{T}_{\theta} \circ \mathcal{C}_{\theta} \left( I_{(i,j)}^{(c)} \right)$. The strategy for down-up-sampling is illustrated in Figure~\ref{fig2:view}.

\textbf{Salience Network:} To mitigate semantic feature distribution overfitting, we employ a salience network, which preserves salience features and forgery patterns. The network comprises five convolutional layers interleaved with three $2 \times 2$ max-pooling layers. Besides, it employs $2 \times 2$ convolutional kernels except for the final $1 \times 1$ projection layer to maintain consistency with the $2 \times 2$ sampling window used in the NPR feature extraction module.
\begin{equation}
  \mathbf{Maps} = SNet\left( |\mathbf{NPR}| \right) \in \mathbb{R}^{H' \times W' \times 64}.
\label{eq:maps}
\end{equation}

\textbf{Top-K Pooling:}
Recall that GAP cannot preferentially focus on the most discriminative local forgery patterns. We propose TKP to select the top-k most discriminative local forgery patterns while deliberately discard these non-discriminative local forgery patterns. However, TKP significantly reduces the difficulty of forged image detection that concurrently introduces a certain degree of overfitting. Thus, we introduce a random deactivation mechanism named RBLD to reduce the model's reliance on the most discriminative local forgery patterns and use RKS to get an auxiliary $\mathbf{Vector}^*$ to avoid overfitting. 
\begin{equation}
   (\mathbf{Vector}, \mathbf{Vector}^*) = TKP(\mathbf{Maps}) \in \mathbb{R}^{2 \times 64k}.
\label{eq:vetor}
\end{equation}

Finally, we classify through a full connection layer. At the same time, build an auxiliary loss by the auxiliary output ($\mathbf{Vector}^*$).
\begin{equation}
(\hat{y}, \hat{y}^*) = \big(\mathcal{FC}_{\theta}(\mathbf{Vector}), \mathcal{FC}_{\theta}(\mathbf{Vector}^*)\big) \in \mathbb{R}^2.
\label{eq:pre_output}
\end{equation}

We adopt the Binary CrossEntropy Loss to construct $\mathcal{L}_{\text{total}}  = \mathcal{L}_{\text{A}} + \alpha \mathcal{L}_{\text{B}} $  ($\alpha=0.1$) to train the deepfake detector. And $X_{train}=\left\{\left(I_{1\left(i,j\right)}^{\left(c\right)},y_1\right),\ldots,\left(I_{n\left(i,j\right)}^{\left(c\right)},y_n\right)\right\}$ was used to optimize the deepfake detector and obtain the best parameter as $\theta_{best}=argmin{(}loss\left(\sigma(LFM\left(\left|\mathbf{NPR}\right|,\theta\right)),y\right)$.
\begin{equation}
    \mathcal{L}_{\text{A}}  = -\frac{1}{N} \sum_{i=1}^N \left[ y_i \log(\hat{y}_i) + (1 - y_i) \log(1 - \hat{y}_i) \right].
\label{eq:lossa}
\end{equation}
\begin{equation}
    \mathcal{L}_{\text{B}}  = -\frac{1}{N} \sum_{i=1}^N \left[y_i \log(\hat{y}_i^{*}) + (1 - y_i) \log(1 - \hat{y}_i^{*}) \right].
\label{eq:lossb}
\end{equation}

\subsection{Regularization}

\begin{algorithm}[h]\small
\caption{TKP with RBLD and RKS}
\label{alg:lfm}
\begin{algorithmic}[1]
\Require Feature maps $\mathbf{Maps} \in \mathbb{R}^{H' \times W' \times 64}$, $k$, $p_\text{min}=0.1$, $p_\text{max}=0.3$.
\Ensure Feature $\mathbf{Vecter} \in \mathbb{R}^{64k}$, $\mathbf{Vector}^* \in \mathbb{R}^{64k}$.
\State Initialize $\mathbf{Vector} \gets \mathbf{0}^{64 \times k}$, $\mathbf{Buffer} \gets [\ ]$ \Comment{Initialisation}
\For{$c \gets 1$ to $64$} \Comment{Process each channel}
    \State Extract feature map: $\mathbf{M}_c \gets \mathbf{Maps}[c,:,:] \in \mathbb{R}^{H'\times W'}$
    \State Sort values: $\{v_{(1)},...,v_{(H'*W')}\} \gets \mathrm{sort}(\mathbf{M}_c)$ \Comment{Ascending order}
    \State Select top-$k$: $\mathcal{T}_k \gets \{v_{(H' * W'-k+1)},...,v_{(H' * W')}\}$

    \For{$i \gets 1$ to $k$} \Comment{Rank-based linear dropout}
        \State $p_i \gets p_\text{min} + (p_\text{max} - p_\text{min}) \times \frac{i-1}{k-1}$ \Comment{Linear prob by rank index}
        \State $r \sim \mathcal{U}(0,1)$
        \If{$r \leq p_i$}
            \State $v_{(H' * W'-k+i)} \gets 0$ \Comment{Apply dropout}
        \EndIf
    \EndFor
    \State $\mathbf{Vecter}[c,:] \gets \mathcal{T}_k$ \Comment{Store processed vector}

    \State $\mathcal{R}_k \gets \text{RandomSample}(\mathbf{Maps}[c,:,:], k)$ \Comment{Random selection}
    \State $\{v_{(1)},...,v_{(k)}\} \gets \mathrm{sort}(\mathcal{R}_k)$ \Comment{Ascending sort}
    \State $\mathbf{Buffer}.\mathrm{append}(\{v_{(1)},...,v_{(k)}\})$ \Comment{Store sorted values}

\EndFor
\State $\mathbf{Vecter} \gets \mathrm{concat}(\mathbf{Vecter})$ \Comment{Flatten into $64k$-dim vector}
\State $\mathbf{Vector}^* \gets \mathrm{concat}(\mathbf{Buffer})$ \Comment{Explicit concatenation} 
\State \Return $\mathbf{Vecter}$, $\mathbf{Vector}^*$
\end{algorithmic}
\end{algorithm}

The proposed Top-K Pooling (TKP) (see Algorithm~\ref{alg:lfm}) preserves the most discriminative top-K local patterns. It makes the training features highly separable, while the decision boundary only classifies well on the training dataset, failing to converge to the optimal position. When transferred to the test set whose features are also separable, the model's generalization drops significantly due to its non-robust decision boundary. To address the problem, we propose two innovations:

\noindent\textbf{(A) Rank-based Linear Dropout}

The TKP algorithm selects and regularizes top-$k$ activations per channel through rank-based linear dropout (see Algorithm~\ref{alg:lfm}). It combines: activation ranking, linear dropout probability scaling ($p=0.1$ to $0.3$ by rank), and channel-wise concatenation that reduces the model's reliance on the most discriminative local forgery patterns.

\noindent\textbf{(B) Random-K Sampling}

The Random-K Sampling module (see Algorithm~\ref{alg:lfm}) complements RBLD through stochastic feature selection. It involves three stages: randomly selecting $k$ activated values, sorting the selected values, and finally concatenating them into a vector of size $64k$. This provides a small amount of gradient paths to non-dominant features and strategic disturbance to enhance model robustness.

\subsection{Inference}
During inference, we first preprocess the input $I_{(i,j)}^{(c)}\in \mathbb{R}^{H\times W\times3}$ into a normalized tensor of size $3 \times 256 \times 256$, then input it into the NPR feature extraction module to obtain initial features. Next, we use SNet to obtain the saliency feature maps of size $64 \times 26 \times 26$. Then, after extracting the top-k discriminative local patterns through TKP, we use a full connection layer to obtain the model output. It is worth noting that to accelerate the inference speed, rank-based linear dropout and random-k sampling are unnecessary. Finally, we use the sigmoid and the decision function to predict the final prediction label.
\section{Experiments}

\subsection{Experimental Setup}

\paragraph{Datasets} To train the deepfake detector, we follow the experimental setting of NPR~\cite{tan2024rethinking} and previous research~\cite{tan2024rethinking,jeong2022bihpf,jeong2022frepgan}  ForenSynths~\cite{wang2020cnn} is employed as the training set. We selected four categories: car, cat, horse, and chair out of the 20 categories in ForenSynths \cite{wang2020cnn} to form the training set. Each category contains 18,000 fake images generated by ProGAN and an equal number of real images from the LSUN~\cite{yu2015lsun} dataset.

We tested a variety of authentic images as well as forged images generated by different GANs and Diffusion Models. The evaluation dataset comprises five primary datasets, including ForenSynths’ dataset~\cite{wang2020cnn}, NPR's GAN data~\cite{tan2024rethinking}, DIRE’s diffusion model dataset~\cite{wang2023dire}, Ojha’s diffusion model dataset~\cite{ojha2023towards}, and NPR’s diffusion model dataset~\cite{tan2024rethinking}, with 38 subdatasets covering images generated by 28 generative models, including ADM~\cite{dhariwal2021diffusion}, DDPM~\cite{ho2020denoising}, IDDPM~\cite{nichol2021improved}, LDM~\cite{rombach2022high}, PNDM~\cite{liu2022pseudo}, VQ-Diffusion~\cite{gu2022vector}, Stable Diffusion v1~\cite{rombach2022high}, and Stable Diffusion v2~\cite{rombach2022high}, Glide~\cite{nichol2021glide}, DALL-E-mini~\cite{ramesh2021zero} and so on. And the real images are sampled from six datasets, including LSUN~\cite{yu2015lsun}, ImageNet~\cite{russakovsky2015imagenet}, CelebA~\cite{liu2015deep}, Celeba-HQ~\cite{karras2017progressive}, COCO~\cite{lin2014microsoft}, and FaceForensics++~\cite{rossler2019faceforensics++}.

\paragraph{Competitors}
We compared the proposed method with several deepfake detection methods, including CNNDetection~\cite{wang2020cnn}, Frank~\cite{frank2020leveraging}, Durall~\cite{durall2020watch}, Patchfor~\cite{chai2020makes}, F3Net~\cite{masi2020two}, SelfBland~\cite{shiohara2022detecting}, GANDetection~\cite{mandelli2022detecting}, BiHPF~\cite{jeong2022bihpf}, FrePGAN~\cite{jeong2022frepgan}, LGrad~\cite{tan2023learning}, Ojha~\cite{ojha2023towards} and NPR~\cite{tan2024rethinking}. 

\subsection{Evaluation of Cross-Domain Generalization}
\begin{table}[htbp]
  \centering
  \scriptsize
  \setlength{\tabcolsep}{0.4pt}
  \begin{tabular}{ccccccccccccccccccccc}
    \toprule
    \multirow{2}{*}{Method} & \multicolumn{2}{c}{AttGAN} & \multicolumn{2}{c}{BEGAN} & \multicolumn{2}{c}{CGAN} & \multicolumn{2}{c}{IMGAN} & \multicolumn{2}{c}{MMDGAN} & \multicolumn{2}{c}{RelGAN} & \multicolumn{2}{c}{S3GAN} & \multicolumn{2}{c}{SNGAN} & \multicolumn{2}{c}{STGAN} & \multicolumn{2}{c}{mean} \\
    \cmidrule(lr){2-3} \cmidrule(lr){4-5} \cmidrule(lr){6-7} \cmidrule(lr){8-9} \cmidrule(lr){10-11} \cmidrule(lr){12-13} \cmidrule(lr){14-15} \cmidrule(lr){16-17} \cmidrule(lr){18-19}
    & {ACC} & {AP} & {ACC} & {AP} & {ACC} & {AP} & {ACC} & {AP} & {ACC} & {AP} & {ACC} & {AP} & {ACC} & {AP} & {ACC} & {AP} & {ACC} & {AP} & {ACC} & {AP} \\
    \midrule
    CNNDetection~\cite{wang2020cnn} & 51.1 & 83.7 & 50.2 & 44.9 & 81.5 & 97.5 & 71.1 & 94.7 & 72.9 & 94.4 & 53.3 & 82.1 & 55.2 & 66.1 & 62.7 & 90.4 & 63.0 & 92.7 & 62.3 & 82.9 \\
    Frank~\cite{frank2020leveraging} & 65.0 & 74.4 & 39.4 & 39.9 & 31.0 & 36.0 & 41.1 & 41.0 & 38.4 & 40.5 & 69.2 & 96.2 & 69.7 & 81.9 & 48.4 & 47.9 & 25.4 & 34.0 & 47.5 & 54.7 \\
    Durall~\cite{durall2020watch} & 39.9 & 38.2 & 48.2 & 30.9 & 60.9 & 67.2 & 50.1 & 51.7 & 59.5 & 65.5 & 80.0 & 88.2 & \second{87.3}    & \second{97.0} & 54.8 & 58.9 & 62.1 & 72.5 & 60.3 & 63.3 \\
    Patchfor~\cite{chai2020makes} & 68.0 & 92.9 & 97.1 & \best{100.0} & 97.8 & \best{99.9} & 93.6 & 98.2 & 97.9 & \best{100.0} & \second{99.6} & \best{100.0} &  66.8 & 68.1 & \second{97.6} & \second{99.8} & 92.7 & 99.8 & 90.1 & 95.4 \\
    F3Net~\cite{masi2020two} & \second{85.2} & 94.8 & 87.1 & 97.5 & 89.5 & \second{99.8} & 67.1 & 83.1 & 73.7 & 99.6 & 98.8 & \best{100.0} & 65.4 & 70.0 & 51.6 & 93.6 & 60.3 & \second{99.9} & 75.4 & 93.1 \\
    SelfBland~\cite{shiohara2022detecting} & 63.1 & 66.1 & 56.4 & 59.0 & 75.1 & 82.4 & 79.0 & 82.5 & 68.6 & 74.0 & 73.6 & 77.8 & 53.2 & 53.9 & 61.6 & 65.0 & 61.2 & 66.7 & 65.8 & 69.7 \\
    GANDetection~\cite{mandelli2022detecting} & 57.4 & 75.1 & 67.9 & \best{100.0} & 67.8 & 99.7 & 67.6 & 92.4 & 67.7 & 99.3 & 60.9 & 86.2 & 69.6 & 83.5 & 66.7 & 90.6 & 69.6 & 97.2 & 66.1 & 91.6 \\
    LGrad~\cite{tan2023learning} & 68.6 & 93.8 & 69.9 & 89.2 & 50.3 & 54.0 & 71.1 & 82.0 & 57.5 & 67.3 & 89.1 & \second{99.1} & 78.5 & 86.0 & 78.0 & 87.4 & 54.8 & 68.0 & 68.6 & 80.8 \\
    Ojha~\cite{ojha2023towards} & 78.5 & \second{98.3} & 72.0 & 98.9 & 77.6 & \second{99.8} & 77.6 & \second{98.9} & 77.6 & \second{99.7} & 78.2 & 98.7 & 85.2 & \best{98.1} & 77.6 & 98.7 & 74.2 & 97.8 & 77.6 & \second{98.8} \\
    NPR~\cite{tan2024rethinking} & 83.0 & 96.2 & \second{99.0} & \second{99.8} & \second{98.7} & 99.0 & \second{94.5} & 98.3 & \second{98.6} & 99.0 & \second{99.6} & \best{100.0} & 79.0 & 80.0 & 88.8 & 97.4 & \second{98.0} & \best{100.0} & \second{93.2} & 96.6 \\
    LFM (ours) & \best{100.0} & \best{100.0} & \best{99.9} & \best{100.0} & \best{98.8} & \best{99.9} & \best{98.5} & \best{99.9} & \best{99.4} & \best{100.0} & \best{100.0} & \best{100.0} & \best{88.4} & 95.5 & \best{99.2} & \best{100.0} & \best{99.9} & \best{100.0} & \best{98.2} & \best{99.5} \\
    \bottomrule
  \end{tabular}
  \caption{Evaluation of cross-domain (GAN-to-GAN) generalization on 9 GANs dataset from  NPR~\cite{tan2024rethinking}. (CGAN and IMGAN denote CramerGAN and InfoMaxGAN, respectively. Best results are in \best{red}, and the second bests are in \second{blue}.}
  \label{tab:Evaluation_of_GAN2}
\end{table}

In this experiment, we evaluate the cross-domain generalization capability of the proposed model: specifically, a fake detector trained on GAN-generated images is tested on a new set of GAN-generated images from unseen sources. The results, presented in Table~\ref{tab:Evaluation_of_GAN2}, show that the proposed LFM model achieves an accuracy of 98.2\%, significantly outperforming NPR’s 93.2\%. This demonstrates LFM's superior ability to generalize across different GAN sources.
Furthermore, the robustness of LFM is corroborated on the ForenSynths dataset~\cite{wang2020cnn}, as shown in Table~\ref{tab:Evaluation_of_GAN1}. Notably, the model is trained using only four classes, while the test data encompasses all classes. This further confirms LFM's strong generalization capability across unseen categories.

\begin{table}[htbp]
  \centering
  \scriptsize
  \setlength{\tabcolsep}{1pt}
  \begin{tabular}{ccccccccccccc}
    \toprule
    \multirow{2}{*}{Method} & \multicolumn{2}{c}{DDPM} & \multicolumn{2}{c}{IDDPM} & \multicolumn{2}{c}{ADM} & \multicolumn{2}{c}{Midjourney} & \multicolumn{2}{c}{DALLE} & \multicolumn{2}{c}{mean} \\
    \cmidrule(lr){2-3} \cmidrule(lr){4-5} \cmidrule(lr){6-7} \cmidrule(lr){8-9} \cmidrule(lr){10-11} \cmidrule(lr){12-13}
    & {ACC} & {AP} & {ACC} & {AP} & {ACC} & {AP} & {ACC} & {AP} & {ACC} & {AP} & {ACC} & {AP} \\
    \midrule
    CNNDetection~\cite{wang2020cnn} & 50.0 & 63.3 & 48.3 & 52.68 & 53.4 & 64.4 & 48.6 & 38.5 & 49.3 & 44.7 & 49.9 & 52.7 \\
    Frank~\cite{frank2020leveraging} & 47.6 & 43.1 & 70.5 & 85.7 & 67.3 & 72.2 & 39.7 & 40.8 & 68.7 & 65.2 & 58.8 & 61.4 \\
    Durall~\cite{durall2020watch} & 54.1 & 53.6 & 63.2 & 71.7 & 39.1 & 40.8 & 45.7 & 47.2 & 53.9 & 52.2 & 51.2 & 53.1 \\
    Patchfor~\cite{chai2020makes} & 54.1 & 66.3 & 35.8 & 34.2 & 68.6 & 73.7 & 66.3 & 68.8 & 60.8 & 65.1 & 57.1 & 61.6 \\
    F3Net~\cite{masi2020two} & 59.4 & 71.9 & 42.2 & 44.7 & 73.4 & 80.3 & 73.2 & 80.4 & 79.6 & 87.3 & 65.5 & 72.9 \\
    SelfBland~\cite{shiohara2022detecting} & 55.3 & 57.7 & 63.5 & 62.5 & 57.1 & 60.1 & 54.3 & 56.4 & 48.8 & 47.4 & 55.8 & 56.8 \\
    GANDetection~\cite{mandelli2022detecting} & 47.3 & 45.5 & 47.9 & 57.0 & 51.0 & 56.1 & 50.0 & 44.7 & 49.8 & 49.7 & 49.2 & 50.6 \\
    LGrad~\cite{tan2023learning} & 59.8 & 88.5 & 45.2 & 46.9 & 72.7 & 79.3 & 68.3 & 76.0 & 75.1 & 80.9 & 64.2 & 74.3 \\
    Ojha~\cite{ojha2023towards} & 69.5 & 80.0 & 64.9 & 74.2 & \second{81.3} & \second{90.8} & 50.0 & 49.8 & 66.3 & 74.6 & 66.4 & 73.9 \\
    NPR~\cite{tan2024rethinking} & \second{88.5} & \best{95.1} & \second{77.9} & \second{84.8} & 75.8 & 79.3 & \second{77.4} & \second{81.9} & \second{80.7} & \second{83.0} & \second{80.1} & \second{84.8} \\
    LFM (ours) & \best{89.4} & \second{92.3} & \best{87.2} & \best{95.6} & \best{89.6} & \best{96.8} & \best{87.5} & \best{96.1} & \best{90.8} & \best{98.1} & \best{88.9} & \best{95.8} \\
    \bottomrule
  \end{tabular}
  \caption{Cross-Diffusion-Sources Evaluation on those Diffusion datasets from NPR~\cite{tan2024rethinking}. (ADM and IDDPM denote guided-diffusion and improved-diffusion, respectively.)}
\label{tab:Evaluation_of_Diffusion3}
\end{table}

We further evaluate the generalization of LFM in cross-domain from GAN-generated images to diffusion model-generated images. In this experiment, three diffusion model datasets are used as testing domain. The results are summarized in Table~\ref{tab:Evaluation_of_Diffusion3}, Table~\ref{tab:Evaluation_of_Diffusion1} and Table~\ref{tab:Evaluation_of_Diffusion2}.  The test results of dataset collected by NPR~\cite{tan2024rethinking} are shown in Table~\ref{tab:Evaluation_of_Diffusion3}. Compared with LGrad and Ojha, the accuracy of LFM improves by 24.7\% and 22.5\%, respectively. And the LFM achieved 88.9\% accuracy compared to 80.1\% for the NPR's ResNet50, and the average accuracy increased from 84.8\% to 95.8\%. Those results demonstrate that LFM's design is useful to enhance the model's ability to capture local forgery patterns, allowing it to generalize effectively from GAN to diffusion model datasets. The diffusion model test data also encompasses diverse categories, which indirectly validates that LFM partially mitigates the cross-category generalization limitations inherent in deepfake detection models.

\subsection{Comparison with State-of-the-art Methods}

\begin{table}[htbp]
  \centering
  \scriptsize
  \setlength{\tabcolsep}{3pt}
  \begin{tabular}{@{}>{\raggedright}p{1.5cm}*{12}{S[table-format=2.1]}cc@{}}
    \toprule
    \multirow{2}{*}{Method} 
    & \multicolumn{2}{c}{Test Set1~\cite{wang2020cnn}} 
    & \multicolumn{2}{c}{Test Set2~\cite{tan2024rethinking}} 
    & \multicolumn{2}{c}{Test Set3~\cite{wang2023dire}} 
    & \multicolumn{2}{c}{Test Set4~\cite{ojha2023towards}} 
    & \multicolumn{2}{c}{Test Set5~\cite{tan2024rethinking}}
    & \multicolumn{2}{c}{Mean} 
    & \multirow{2}{*}{FLOPs} 
    & \multirow{2}{*}{Params} \\
    \cmidrule(lr){2-3} \cmidrule(lr){4-5} \cmidrule(lr){6-7} \cmidrule(lr){8-9} \cmidrule(lr){10-11} \cmidrule(lr){12-13}
    & {ACC} & {AP} & {ACC} & {AP} & {ACC} & {AP} & {ACC} & {AP} & {ACC} & {AP} & {ACC} & {AP} & & \\
    \midrule
    LGrad~\cite{tan2023learning} & 86.1 & 91.5 & 68.6 & 80.8 & 88.2 & 98.5 & 90.9 & 97.2 & 74.3 & 64.2 & 80.5 & 88.5 & 50.95B & 48.61M \\
    Ojha~\cite{ojha2023towards} & 89.1 & 98.3 & 77.6 & 98.8 & 74.1 & 91.7 & 86.6 & 94.5 & 66.4 & 73.9 & 79.8 & 91.4 & 51.90B & 202.05M \\
    NPR~\cite{tan2024rethinking} & 92.5 & 96.1 & 93.2 & 96.6 & 95.3 & \second{99.8} & \second{95.2} & 97.4 & 80.1 & 84.8 & 92.2 & 95.8 &  \second{2.30B} &  \second{1.44M} \\
    Fatformer~\cite{liu2024forgery} & \best{98.3} & \best{99.7} &  \second{98.9} & \best{100.0} & 93.8 & 98.5 & 72.5 & 82.2 & 66.8 & 85.4 & 87.8 & 93.7 & 577.25B & 577.25M \\
    SAFE~\cite{li2024improving} &  \second{96.2} &  \second{98.8} & \best{99.0} & \second{99.8} &  \second{95.7} & 99.1 & 91.2 & \best{99.0} &  \second{81.9} &  \second{91.8} &  \second{93.8} &  \second{98.2} &  \second{2.30B} &  \second{1.44M} \\
    LFM (ours) & 95.5 & 98.3 & 98.2 & 99.5 & \best{98.7} & \best{100.0} & \best{95.4} & \second{98.4} & \best{88.9} & \best{95.8} & \best{95.9} & \best{98.6} & \best{1.61B} & \best{0.72M} \\
    \bottomrule
  \end{tabular}
  \caption{The evaluation results of the most recent and effective deepfake detection methods on these 38 datasets. (The results of Fatformer~\cite{liu2024forgery} and SAFE~\cite{li2024improving} were tested on the datasets we used based on the best weights provided by the official sources.)}
  \label{tab:overall_evaluation}
\end{table}

We averaged the test results of 38 sub-datasets as shown in Table~\ref{tab:overall_evaluation}. Improving our classification model from ResNet50 to LFM can increase accuracy from 92.2\% to 95.9\% and average precision from 95.8\% to 98.6\%. And the accuracy is over 2.1\% than the SAFE~\cite{li2024improving}. This sufficiently demonstrates that the local focus mechanism we proposed can enhance the model's ability to capture forgery patterns, thereby strengthening its generalization capability. In addition, LFM has a significant advantage in detection speed at 1789 FPS with the A6000.

\subsection{Ablation Studies}

\begin{table}[hbt]
\centering
\begin{tabular}{lcccccc}
\hline
\textbf{Configuration} & \textbf{SNet} & \textbf{TKP-base} & \textbf{RBLD} & \textbf{RKS} & \textbf{ACC} & $\Delta \textbf{ACC}$ \\ \hline
NPR~\cite{tan2024rethinking}      &  &  &  &  & 90.6 & - \\ 
+SNet         & $\checkmark$ &  &  &  & 92.4 & +1.8 \\
+TKP-base     & $\checkmark$ & $\checkmark$ &  &  & 94.5 & +1.9 \\
+RBLD          & $\checkmark$ & $\checkmark$ & $\checkmark$ &  & 95.5 & +1.0 \\
\hline
+RKS(\textbf{Full}) & $\checkmark$ & $\checkmark$ & $\checkmark$ & $\checkmark$ & \textbf{95.9} & \textbf{+0.4} \\ \hline
\end{tabular}
\caption{Component-wise ablation analysis (\%)}
\label{tab:ablation}
\end{table}

As shown in Table~\ref{tab:ablation}, the NPR~\cite{tan2024rethinking} model achieves 90.6\% accuracy, while our full LFM (SNet+TKP-base+RBLD+RKS) reaches 95.9\%. The SNet delivered a 1.8\% improvement, demonstrating that the network is simple yet effective. The TKP-base module improves model by 1.9\%, highlighting the benefit of focusing on local patterns over GAP. The RBLD module contributes an additional 1.0\% gain, demonstrating its effectiveness in suppressing overfitting. And the RKS module provides a modest improvement of 0.4\%. The full LFM achieves a 5.3\% absolute improvement, confirming the synergistic effect of combining these modules.

\subsection{Visualization}
\label{subsec:visualization}

\begin{figure}[t]
    \centering
    \includegraphics[width=1.0\linewidth]{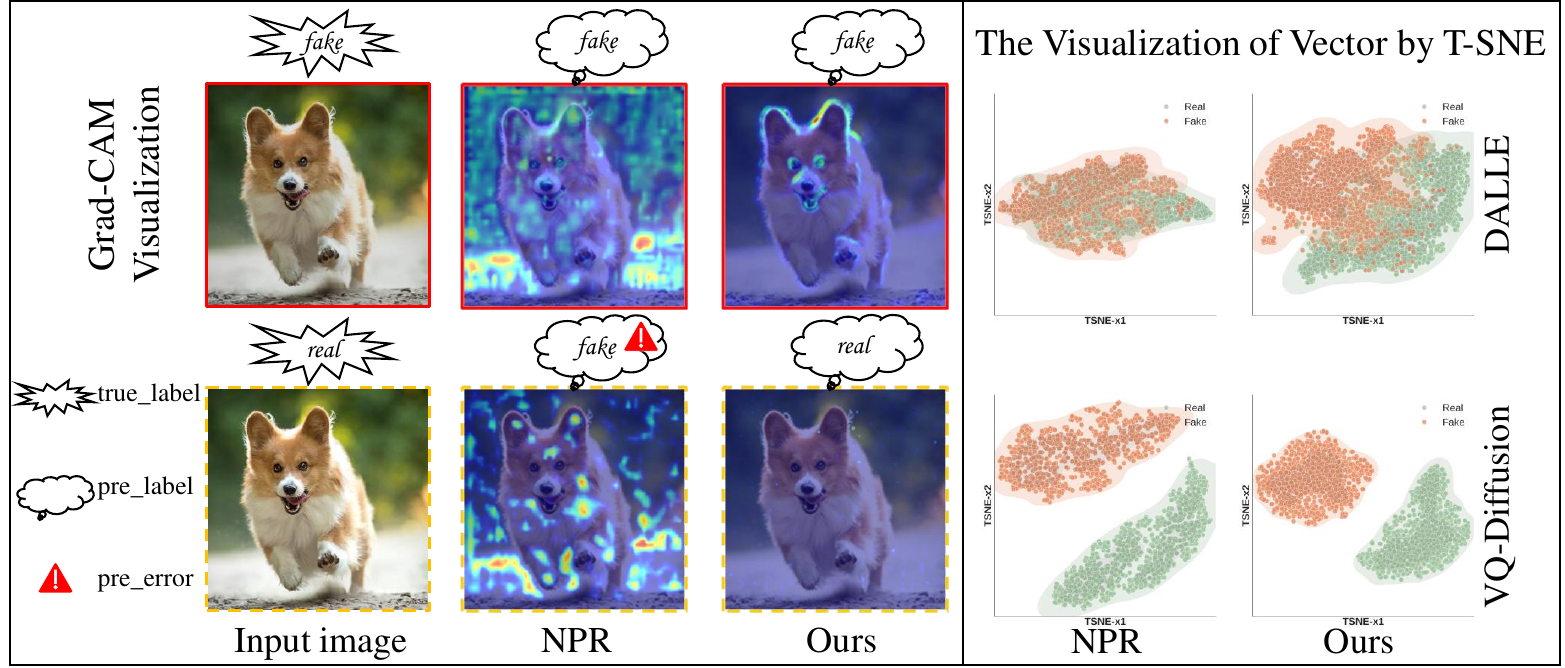}
    \caption{The visualizations highlight the differences between our approach and NPR~\cite{tan2024rethinking}. On the left, Grad-CAM~\cite{selvaraju2016grad} is used to visualize attention maps, while on the right, T-SNE~\cite{van2008visualizing} is applied to dataset embeddings, including samples from DALLE and VQ-Diffusion.}
    \label{fig6:visualization}
\end{figure}

We also conducted visual analytics, as illustrated in Figure~\ref{fig6:visualization}. First, we applied Grad-CAM~\cite{selvaraju2016grad} to highlight the key regions the models attend to during classification. The resulting heatmaps show that our proposed LFM method focuses more accurately on localized forgery artifacts, which are critical for deepfake detection. In contrast, NPR exhibits a tendency to attend to broader, less relevant regions. This targeted focus allows LFM to achieve greater discriminative power between real and fake images. Additionally, we employed T-SNE~\cite{van2008visualizing} to reduce the dimensionality of the feature vectors input to the fully connected layer for 2D visualization. On the DALLE dataset, the features extracted using TKP demonstrate superior linear separability compared to those from GAP. While both methods achieve linear separability on the VQ-Diffusion dataset, TKP produces features with tighter intra-class clustering and greater inter-class separation, indicating more robust representational capacity. 
\section{Conclusion}
Our work reconsiders the applicability of traditional networks in deepfake detection tasks based on reconstruction learning. In order to solve the problem that deep networks are vulnerable to semantic interference in deepfake detection tasks and the lack of local feature sensitivity of traditional pooling strategies, we propose a simple yet effective local focusing mechanism combining salience networks, which can effectively enhance the model's ability to capture forged patterns. Experiments show that our method achieves excellent detection results on many test sets. Meanwhile, thanks to a simple but effective network design and pooling strategy, our method achieves a significant advantage in detection speed at 1789 FPS with the A6000. In addition, our method can be combined with methods such as NPR~\cite{tan2024rethinking}, reconstructed differential features~\cite{cao2022end}, SAFE~\cite{li2024improving}, etc.

\section{Acknowledgment}
This project was partially supported by NSFC 62372150.

\bibliography{egbib}

\appendix
\section*{Appendix}
\setcounter{figure}{0}
\setcounter{equation}{0}
\renewcommand{\thetable}{\Roman{table}}
\renewcommand{\thefigure}{\Roman{figure}}
\renewcommand{\theequation}{\Roman{equation}}

\section{Implementation Details}
In feature extraction of NPR~\cite{tan2024rethinking} , the down-sampling window is set to $2 \times 2$, and parameter $j^{*}$ is set to 0. In addition, we take into account that the NPR down-up-sampling window is 2. Thus, we use a $ 2\times 2$ convolution instead of the common $3 \times 3$ convolution in SNet. The experiment was conducted under the PyTorch 2.6.0 framework, using a single NVIDIA A6000 GPU for training and testing. The initial learning rate is set to $1 \times 10^{-4}$, the batch size is 32, and the optimizer uses Adam. The evaluation metrics and strategies were consistent with the NPR~\cite{tan2024rethinking} approach, using accuracy (ACC) and average accuracy (AP) as the primary performance evaluation metrics. In addition, we added FPS, FLOPs to consider the inference speed of the model and Params to measure the complexity of the model.

\section{Study on The Number of Convolution Layers}\label{app:layer_num_comparison}
\begin{table}[h]
  \centering
  \begin{tabular}{ccccccc}
    \toprule
    \multirow{2}{*}{Method} & \multicolumn{5}{c}{SNet(Number of Layers)} & \multirow{2}{*}{ResNet50} \\
    \cmidrule(lr){2-6}
    & 4 & 5 & 6 & 7 & 8 & \\
    \midrule
    ACC (\%) & 93.8 & \best{95.9} & 94.9 & 94.5 & 94.7 & 93.0 \\
    AP (\%) & 95.0 & \best{98.6} & 97.8 & 97.9 & 97.6 & 97.1 \\
    \bottomrule
  \end{tabular}
  \caption{Experimental results of TKP combined with different convolutional neural networks}
  \label{tab:layer_num_comparison}
\end{table}

Considering that fewer convolutional layers tend to suppress semantics more effectively, but they also result in weaker fitting capabilities. Therefore, we focus on exploring the impact mechanism of convolution layers on deepfake detection performance. Experimental results show that model performance is significantly sensitive to the number of convolution layers: this indicates that salience networks need to balance representation capability with semantic interference, and a 4-layer architecture achieves optimal balance in this task. 

\section{Comparative Analysis on Different Pooling Methods}\label{app:module_comparison}

\begin{table}[h]
  \centering
  \begin{tabular}{cccc}
    \toprule
    \textbf{module} & \textbf{ACC} & \textbf{AP} & \textbf{Parameters} \\
    \midrule
    GAP       & 92.4 & 94.1 & 0 \\
    GMP       & 95.0 & 98.1 & 0 \\
    Attention & 95.1 & 97.7 & $\checkmark$ \\ 
    TKP       & \best{95.9} & \best{98.6} & 0 \\
    \bottomrule
  \end{tabular}
  \caption{Experimental metrics for constructing deepfake detection models using GAP, GMP, TKP and Attention in combination with SNet respectively. (The Attention module is designed based on the spatial attention module in CBAM~\cite{woo2018cbam}.)}
  \label{tab:module_comparison}
\end{table}

To further explore the effectiveness of the TKP pooling strategy, we conducted experiments on the SNet architecture using GAP, GMP, TKP, and Attention, respectively. The experimental results are shown in the above table. It can be seen that compared with the traditional GAP, using GMP can effectively improve the performance of the model (95.0\% vs 92.4\%) due to only focusing on top-1 discriminative local forgery patterns. Moreover, TKP addresses the drawback of GMP being too arbitrary. Additionally, TKP introduces a random deactivation mechanism and random-K sampling during the training stage to avoid overfitting, further enhancing the model's performance (95.9\% vs 95.0\%). On the other hand, we compared it with the Attention mechanism. The experiments show that the TKP with zero parameters we designed can still achieve better detection results (95.9\% vs 95.1\%) than Attention.

\section{Sensitivity Analysis of Hyperparameter k}\label{app:k_comparison}
\begin{table}[h]
  \centering
  \begin{tabular}{c *{5}{S[table-format=3.1]}}
    \toprule
    {$k$} & {1} & {4} & {16} & {64} & {256} \\
    \midrule
    ACC (\%) & 94.8 & 95.4 & 95.9 & 95.8 & 95.8 \\
    AP (\%)  & 97.4 & 98.1 & 98.6 & 97.9 & 98.1 \\
    \bottomrule
  \end{tabular}
  \caption{The model performance is influenced by different values of the hyperparameter $k$}
  \label{tab:k_sensitivity}
\end{table}

The selection of hyperparameter $k$ in the Top-k feature selection process significantly impacts model performance. As shown in Table~\ref{tab:k_sensitivity}, when $k$ increases from 1 to 16, the model's ACC improves from 94.8\% to 95.9\%, while the AP increases from 97.4\% to 98.6\%. This demonstrates that appropriately expanding the number of selected features helps capture more comprehensive discriminative information. However, when k continues to increase beyond 16 up to 256, ACC stabilizes at 95.8\%-95.9\%, while AP exhibits fluctuations (dropping to 97.9\% at $k=64$). This indicates that excessively large k values lead to feature redundancy, which not only fails to further improve performance but may also introduce noise. The experimental results demonstrate that $k=16$ represents the optimal choice for this task, achieving the best balance between computational efficiency and feature discriminative power.

\section{More Results of Cross-Domain Generalization}

\begin{table}[htbp]
  \centering
  \scriptsize
  \setlength{\tabcolsep}{1pt}
  \begin{tabular}{ccccccccccccccccccc}
    \toprule
    \multirow{2}{*}{Method} & \multicolumn{2}{c}{progan} & \multicolumn{2}{c}{stylegan} & \multicolumn{2}{c}{stylegan2} & \multicolumn{2}{c}{biggan} & \multicolumn{2}{c}{cyclegan} & \multicolumn{2}{c}{Stargan} & \multicolumn{2}{c}{guagan} & \multicolumn{2}{c}{deepfake} & \multicolumn{2}{c}{mean} \\
    \cmidrule(lr){2-3} \cmidrule(lr){4-5} \cmidrule(lr){6-7} \cmidrule(lr){8-9} \cmidrule(lr){10-11} \cmidrule(lr){12-13} \cmidrule(lr){14-15} \cmidrule(lr){16-17} \cmidrule(lr){18-19}
    & {ACC} & {AP} & {ACC} & {AP} & {ACC} & {AP} & {ACC} & {AP} & {ACC} & {AP} & {ACC} & {AP} & {ACC} & {AP} & {ACC} & {AP} & {ACC} & {AP} \\
    \midrule
    CNNDetection~\cite{wang2020cnn} & 91.4 & 99.4 & 63.8 & 91.4 & 76.4 & 97.5 & 52.9 & 73.3 & 72.7 & 88.6 & 63.8 & 90.8 & 63.9 & 92.2 & 51.7 & 62.3 & 67.1 & 86.9 \\
    Frank~\cite{frank2020leveraging} & 90.3 & 85.2 & 74.5 & 72.0 & 73.1 & 71.4 & 88.7 & 86.0 & 75.5 & 71.2 & 99.5 & 99.5 & 69.2 & 77.4 & 60.7 & 49.1 & 78.9 & 76.5 \\
    Durall~\cite{durall2020watch} & 81.1 & 74.4 & 54.4 & 52.6 & 66.8 & 62.0 & 60.1 & 56.3 & 69.0 & 64.0 & 98.1 & 98.1 & 61.9 & 57.4 & 50.2 & 50.0 & 67.7 & 64.4 \\
    Patchfor~\cite{chai2020makes} & 97.8 & \best{100.0} & 82.6 & 93.1 & 83.6 & 98.5 & 64.7 & 69.5 & 74.5 & 87.2 & \best{100.0} & \best{100.0} & 57.2 & 55.4 & 85.0 & 93.2 & 80.7 & 87.1 \\
    F3Net~\cite{masi2020two} & 99.4 & \best{100.0} & 92.6 & 99.7 & 88.0 & 99.8 & 65.3 & 69.9 & 76.4 & 84.3 & \best{100.0} & \best{100.0} & 58.1 & 56.7 & 63.5 & 78.8 & 80.4 & 86.2 \\
    SelfBland~\cite{shiohara2022detecting} & 58.8 & 65.2 & 50.1 & 47.7 & 48.6 & 47.4 & 51.1 & 51.9 & 59.2 & 65.3 & 74.5 & 89.2 & 59.2 & 65.5 & \best{93.8} & \best{99.3} & 61.9 & 66.4 \\
    GANDetection~\cite{mandelli2022detecting} & 82.7 & 95.1 & 74.4 & 92.9 & 69.9 & 87.9 & 76.3 & 89.9 & 85.2 & 95.5 & 68.8 & 99.7 & 61.4 & 75.8 & 60.0 & 83.9 & 72.3 & 90.1 \\
    BiHPF~\cite{jeong2022bihpf} & 90.7 & 86.2 & 76.9 & 75.1 & 76.2 & 74.7 & 84.9 & 81.7 & 81.9 & 78.9 & 94.4 & 94.4 & 69.5 & 78.1 & 54.4 & 54.6 & 78.6 & 77.9 \\
    FrePGAN~\cite{jeong2022frepgan} & 99.0 & \second{99.9} & 80.7 & 89.6 & 84.1 & 98.6 & 69.2 & 71.1 & 71.1 & 74.4 & \second{99.9} & \best{100.0} & 60.3 & 71.7 & 70.9 & 91.9 & 79.4 & 87.2 \\
    LGrad~\cite{tan2023learning} & \best{99.9} & \best{100.0} & 94.8 & \second{99.9} & 96.0 & \second{99.9} & 82.9 & 90.7 & 85.3 & 94.0 & 99.6 & \best{100.0} & 72.4 & 79.3 & 58.0 & 67.9 & 86.1 & 91.5 \\
    Ojha~\cite{ojha2023towards} & 99.7 & \best{100.0} & 89.0 & 98.7 & 83.9 & 98.4 & \second{90.5} & \best{99.1} & 87.9 & \best{99.8} & 91.4 & \best{100.0} & \best{89.9} & \best{100.0} & 80.2 & 90.2 & 89.1 & \best{98.3} \\
    NPR~\cite{tan2024rethinking} & \second{99.8} & \best{100.0} & \second{96.3} & 99.8 & \second{97.3} & \best{100.0} & 87.5 & 94.5 & \best{95.0} & 99.5 & 99.7 & \best{100.0} & 86.6 & 88.8 & 77.4 & 86.2 & \second{92.5} & \second{96.1} \\
    LFM (Ours) & \second{99.8} & \best{100.0} & \best{99.9} & \best{100.0} & \best{99.9} & \best{100.0} & \best{91.2} & \second{97.6} & \second{94.3} & \second{99.6} & \best{100.0} & \best{100.0} & \second{88.3} & \second{93.0} & \second{90.8} & \second{95.8} & \best{95.5} & \best{98.3} \\
    \bottomrule
  \end{tabular}
  \caption{Evaluation on GAN-to-GAN cross-domain generalization. The test set is ForenSynths~\cite{wang2020cnn}. All the results except those of LFM are from NPR~\cite{tan2024rethinking}.}
  \label{tab:Evaluation_of_GAN1}
  \end{table}

\begin{table}[htbp]
  \centering
  \scriptsize
  \setlength{\tabcolsep}{1pt}
  \begin{tabular}{ccccccccccccccccccc}
    \toprule
    \multirow{2}{*}{Method} & \multicolumn{2}{c}{ADM} & \multicolumn{2}{c}{DDPM} & \multicolumn{2}{c}{IDDPM} & \multicolumn{2}{c}{LDM} & \multicolumn{2}{c}{PNDM} & \multicolumn{2}{c}{VQ-Diffusion} & \multicolumn{2}{c}{SD-v1} & \multicolumn{2}{c}{SD-v2} & \multicolumn{2}{c}{mean} \\
    \cmidrule(lr){2-3} \cmidrule(lr){4-5} \cmidrule(lr){6-7} \cmidrule(lr){8-9} \cmidrule(lr){10-11} \cmidrule(lr){12-13} \cmidrule(lr){14-15} \cmidrule(lr){16-17} \cmidrule(lr){18-19}
    & {ACC} & {AP} & {ACC} & {AP} & {ACC} & {AP} & {ACC} & {AP} & {ACC} & {AP} & {ACC} & {AP} & {ACC} & {AP} & {ACC} & {AP} & {ACC} & {AP} \\
    \midrule
    CNNDetection~\cite{wang2020cnn} & 53.9 & 71.8 & 62.7 & 76.6 & 50.2 & 82.7 & 50.4 & 78.7 & 50.8 & 90.3 & 50.0 & 71.0 & 38.0 & 76.7 & 52.0 & 90.3 & 51.0 & 79.8 \\
    Frank~\cite{frank2020leveraging} & 58.9 & 65.9 & 37.0 & 27.6 & 51.4 & 65.0 & 51.7 & 48.5 & 44.0 & 38.2 & 51.7 & 66.7 & 32.8 & 52.3 & 40.8 & 37.5 & 46.0 & 50.2 \\
    Durall~\cite{durall2020watch} & 39.8 & 42.1 & 52.9 & 49.8 & 55.3 & 56.7 & 43.1 & 39.9 & 44.5 & 47.3 & 38.6 & 38.3 & 39.5 & 56.3 & 62.1 & 55.8 & 47.0 & 48.3 \\
    Patchfor~\cite{chai2020makes} & 77.5 & 93.9 & 62.3 & 97.1 & 50.0 & 91.6 & 99.5 & \best{100.0} & 50.2 & \second{99.9} & \best{100.0} & \best{100.0} & 90.7 & \second{99.8} & 94.8 & \best{100.0} & 78.1 & 97.8 \\
    F3Net~\cite{masi2020two} & 80.9 & 96.9 & 84.7 & \second{99.4} & 74.7 & 98.9 & \best{100.0} & \best{100.0} & 72.8 & 99.5 & \best{100.0} & \best{100.0} & 73.4 & 97.2 & \best{99.8} & \best{100.0} & 85.8 & 99.0 \\
    SelfBland~\cite{shiohara2022detecting} & 57.0 & 59.0 & 61.9 & 49.6 & 63.2 & 66.9 & 83.3 & 92.2 & 48.2 & 48.2 & 77.2 & 82.7 & 46.2 & 68.0 & 71.2 & 73.9 & 63.5 & 67.6 \\
    GANDetection~\cite{mandelli2022detecting} & 51.1 & 53.1 & 62.3 & 46.4 & 50.2 & 63.0 & 51.6 & 48.1 & 50.6 & 79.0 & 51.1 & 51.2 & 39.8 & 65.6 & 50.1 & 36.9 & 50.8 & 55.4 \\
    LGrad~\cite{tan2023learning} & 86.4 & 97.5 & \best{99.9} & \best{100.0} & 66.1 & 92.8 & \second{99.7} & \best{100.0} & 69.5 & 98.5 & \second{96.2} & \best{100.0} & 90.4 & 99.4 & \second{97.1} & \best{100.0} & 88.2 & 98.5 \\
    Ojha~\cite{ojha2023towards} & 78.4 & 92.1 & 72.9 & 78.8 & 75.0 & 92.8 & 82.2 & \second{97.1} & 75.3 & 92.5 & 83.5 & \second{97.7} & 56.4 & 90.4 & 71.5 & \second{92.4} & 74.4 & 91.7 \\
    NPR~\cite{tan2024rethinking} & \second{88.6} & \second{98.9} & \second{99.8} & \best{100.0} & \second{91.8} & \second{99.8} & \best{100.0} & \best{100.0} & \second{91.2} & \best{100.0} & \best{100.0} & \best{100.0} & \best{97.4} & \second{99.8} & 93.8 & \best{100.0} & \second{95.3} & \second{99.8} \\
    LFM (ours) & \best{96.5} & \best{99.8} & \best{99.9} & \best{100.0} & \best{97.9} & \best{100.0} & \best{100.0} & \best{100.0} & \best{99.4} & \best{100.0} & \best{100.0} & \best{100.0} & \second{96.2} & \best{99.9} & \best{99.8} & \best{100.0} & \best{98.7} & \best{100.0} \\
    \bottomrule
  \end{tabular}
  \caption{Evaluation on Diffusion-to-Diffusion cross-domain generalization. The test set is DiffusionForensics~\cite{wang2023dire}. (ADM, SD-v1 and SD-v2 denote guided, Stable Diffusion v1 and Stable Diffusion v2, respectively.)}
  \label{tab:Evaluation_of_Diffusion1}
\end{table}

\begin{table}[htbp]
  \centering
  \scriptsize
  \setlength{\tabcolsep}{1pt}
  \begin{tabular}{ccccccccccccccccccc}
    \toprule
    \multirow{2}{*}{Method} & \multicolumn{2}{c}{DALLE} & \multicolumn{2}{c}{G\_100\_10} & \multicolumn{2}{c}{G\_100\_27} & \multicolumn{2}{c}{G\_50\_27} & \multicolumn{2}{c}{ADM} & \multicolumn{2}{c}{L\_100} & \multicolumn{2}{c}{L\_200} & \multicolumn{2}{c}{L\_200\_cfg} & \multicolumn{2}{c}{mean} \\
    \cmidrule(lr){2-3} \cmidrule(lr){4-5} \cmidrule(lr){6-7} \cmidrule(lr){8-9} \cmidrule(lr){10-11} \cmidrule(lr){12-13} \cmidrule(lr){14-15} \cmidrule(lr){16-17} \cmidrule(lr){18-19}
    & {ACC} & {AP} & {ACC} & {AP} & {ACC} & {AP} & {ACC} & {AP} & {ACC} & {AP} & {ACC} & {AP} & {ACC} & {AP} & {ACC} & {AP} & {ACC} & {AP} \\
    \midrule
    CNNDetection~\cite{wang2020cnn} & 51.8 & 61.3 & 53.3 & 72.9 & 53.0 & 71.3 & 54.2 & 76.0 & 54.9 & 66.6 & 51.9 & 63.7 & 52.0 & 64.5 & 51.6 & 63.1 & 52.8 & 67.4 \\
    Frank~\cite{frank2020leveraging} & 57.0 & 62.5 & 53.6 & 44.3 & 50.4 & 40.8 & 52.0 & 42.3 & 53.4 & 52.5 & 56.6 & 51.3 & 56.4 & 50.9 & 56.5 & 52.1 & 54.5 & 49.6 \\
    Durall~\cite{durall2020watch} & 55.9 & 58.0 & 54.9 & 52.3 & 48.9 & 46.9 & 51.7 & 49.9 & 40.6 & 42.3 & 62.0 & 62.6 & 61.7 & 61.7 & 58.4 & 58.5 & 54.3 & 54.0 \\
    Patchfor~\cite{chai2020makes} & 79.8 & \second{99.1} & 87.3 & 99.7 & 82.8 & 99.1 & 84.9 & \second{98.8} & 74.2 & 81.4 & 95.8 & 99.8 & 95.6 & \second{99.9} & 94.0 & \second{99.8} & 86.8 & 97.2 \\
    F3Net~\cite{masi2020two} & 71.6 & 79.9 & 88.3 & 95.4 & 87.0 & 94.5 & 88.5 & 95.4 & 69.2 & 70.8 & 74.1 & 84.0 & 73.4 & 83.3 & 80.7 & 89.1 & 79.1 & 86.5 \\
    SelfBland~\cite{shiohara2022detecting} & 52.4 & 51.6 & 58.8 & 63.2 & 59.4 & 64.1 & 64.2 & 68.3 & 58.3 & 63.4 & 53.0 & 54.0 & 52.6 & 51.9 & 51.9 & 52.6 & 56.3 & 58.7 \\
    GANDetection~\cite{mandelli2022detecting} & 67.2 & 83.0 & 51.2 & 52.6 & 51.1 & 51.9 & 51.7 & 53.5 & 49.6 & 49.0 & 54.7 & 65.8 & 54.9 & 65.9 & 53.8 & 58.9 & 54.3 & 60.1 \\
    LGrad~\cite{tan2023learning} & 88.5 & 97.3 & 89.4 & 94.9 & 87.4 & 93.2 & 90.7 & 95.1 & \second{86.6} & \best{100.0} & 94.8 & 99.2 & 94.2 & 99.1 & 95.9 & 99.2 & 90.9 & 97.2 \\
    Ojha~\cite{ojha2023towards} & \second{89.5} & 96.8 & 90.1 & 97.0 & 90.7 & 97.2 & 91.1 & 97.4 & 75.7 & 85.1 & 90.5 & 97.0 & 90.2 & 97.1 & 77.3 & 88.6 & 86.9 & 94.5 \\
    NPR~\cite{tan2024rethinking} & \best{94.5} & \best{99.5} & \second{98.2} & \second{99.8} & \second{97.8} & \second{99.7} & \second{98.2} & \second{99.8} & 75.8 & 81.0 & \second{99.3} & \second{99.9} & \second{99.1} & \second{99.9} & \second{99.0} & \second{99.8} & \second{95.2} & \second{97.4} \\
    LFM (ours) & 78.6 & 94.7 & \best{99.8} & \best{100.0} & \best{99.5} & \best{100.0} & \best{99.7} & \best{100.0} & \best{86.9} & \second{92.4} & \best{99.7} & \best{100.0} & \best{99.7} & \best{100.0} & \best{99.5} & \best{99.9} & \best{95.4} & \best{98.4} \\
    \bottomrule
  \end{tabular}
  \caption{Evaluation on GAN-to-GAN cross-domain generalization. The test set is Ojha~\cite{ojha2023towards}. (G and L denote Glide and LDM, respectively)}
  \label{tab:Evaluation_of_Diffusion2}
\end{table}

\end{document}